# From Organisational Structure to Organisational Behaviour Formalisation[*]

Catholijn M. Jonker[1] and Jan Treur[2]

[1]Delft University of Technology, Department of Man-Machine Interaction
Mekelweg 4, NL-2628 CD Delft, The Netherlands
         Email: c.m.jonker@tudelft.nl
         URL: http://www.mmi.tudelft.nl/
[2]Vrije Universiteit Amsterdam, Department of Artificial Intelligence
De Boelelaan 1081a, NL-1081 HV Amsterdam, The Netherlands
                Email: treur@cs.vu.nl
         URL: http://www.cs.vu.nl/~treur

**Abstract**  To understand how an organisational structure relates to organisational behaviour is an interesting fundamental challenge in the area of organisation modelling. Specifications of organisational structure usually have a diagrammatic form that abstracts from more detailed dynamics. Dynamic properties of agent systems, on the other hand, are often specified in the form of a set of logical formulae in some temporal language. This paper addresses the question how these two perspectives can be combined in one framework. It is shown how for different aggregation levels and other elements within an organisation structure, sets of dynamic properties can be specified. Organisational structure provides a structure of (interlevel) relationships between these multiple sets of dynamic properties. Thus organisational structure is reflected in the formalisation of the dynamics of organisational behaviour. To illustrate the effectiveness of the approach a formal foundation is presented for the integrated specification of both structure and behaviour of an AGR organisation model.

## 1 Introduction

As the complexity of systems based on multiple software agents increases, as is the case, for example in the context of Internet, their dynamics are less easy to predict and to manage. A recent development is to incorporate organisation modelling methods within the software engineering process of multi-agent systems. Indeed, like complex agent-based software systems, societies are characterised by complex dynamics involving interaction between large numbers of actors and groups of actors. If within society such complex dynamics would take place in an completely unstructured, incoherent manner, any actor involved has not much to rely on to do prediction, and therefore is not able to function in a knowledgeable manner. This has serious disadvantages, which is a reason why in history within human societies organisational structure has been developed as a means to manage complex dynamics. Here it is assumed that organisational structure provides co-ordination of the processes in such a manner that a process or agent involved can function in a more adequate manner. So the basic assumption is that providing organisational structure has implications to organisational dynamics. The dynamics induced by a given organisational structure are much more dependable than in an entirely unstructured situation. It is assumed that the organisational structure itself is relatively stable, i.e., the structure may change, but the frequency and scale of change are

---

assumed low compared to the more standard dynamics through the structure. Within the field of Organisation Theory such organisational structures regulating societal dynamics are studied; e.g., [29], [37]. In this literature, indeed the basic assumption is that organisational structure is used to obtain effective and efficient, competitive dynamics (or organisational behaviour) of a desired type. However, a main issue, not addressed in more detail in this largely informal and often guru-like literature in Organisation Theory is:

> *How exactly does such organisational structure relate to organisational dynamics*?

This question is the main focus of this paper. For further analysis a crucial issue here is how exactly a formalisation of organisational structure can be obtained and how that formalisation can be related to a formalisation of organisational dynamics. This implies in particular that as a foundation, both the structural aspects and the dynamic aspects have to be formalised in an appropriate manner and doing so their relation should be clarified. To address this from a generic foundational metamodelling perspective is the aim of the paper. A formalised approach is presented that can be used as a metamodel to be applied to any specific organisation modelling approach and dynamic modelling approach. As a case study it is shown how this metamodel applies to the organisation modelling approach AGR (cf. [12]) in combination with a dynamic modelling approach, for example, based on TTL (cf. [28]).

In this paper first in Section 2 developments in organisation modelling are discussed, and in particular the AGR approach that is used as an organisational modelling approach case study here. Next, it is explored in more detail (in Section 3) how *organisational structure* can be specified based on a formally defined foundation. Section 4 addresses how, given a formalisation of organisational structure, *sets of dynamic properties* can be associated to each aggregation level and element within this structure. These dynamic properties can be used for simulation (especially when expressed in executable format) and formal analysis of empirical or simulated traces.

As different parts or aggregation levels are structurally related within an organisational structure, a next question in the context of the relation between organisation structure and dynamics is how the associated sets of dynamic properties can be related accordingly. To this end, in Section 5, as part of an analysis of how organisational structure relates to organisational dynamics, *logical interlevel relationships between sets of dynamic properties* of different elements or aggregation levels within an organisational structure are described. Finally, for realisation of an organisation, requirements can be specified on *agents allocated to roles* within an organisation model (Section 6). This provides a generic metamodel that can be used as a foundation for integrated specification languages covering both structure and dynamics. However, the specific choice of organisation modelling and dynamic modelling languages (for example, also using semi-formal and graphical elements) is left open. As an illustration two example dynamic modelling languages are discussed: TTL (Temporal Trace Language) and an extension of LTL (Linear Time Logic). As a case study the introduced metamodel is illustrated for the AGR organisation modelling approach, but it has a wider applicability, as is discussed in Section 7.

## 2  Developments in Organisation Modelling

In this section, first in Section 2.1 an overview of organisation modelling literature is presented, and next, in Section 2.2 one approach is described in some more detail: the AGR approach (cf. [12]).

## 2.1 Overview

Multi-agent or organisation modelling approaches have been developed in three manners. Firstly, informal or semi-formal graphical representations of the organisational structure have been developed; i.e., pictures with boxes and arrows; e.g., [37]. Such organisation models, although they provide a detailed account of the organisation structure, remain on a rather abstract level. In particular they do not give indications how the more detailed dynamics takes place; it does not specify how these structures relate to dynamics.

Secondly, within the area of Computational Organisation Theory and Artificial Intelligence, a number of organisation modelling approaches have been developed to simulate and analyse dynamics within organisations in society; e.g., [39], [32], [38], [12], [16], [23], [26]. Some of these approaches explicitly focus on modelling organisational structure, abstracting from the detailed dynamics. Other approaches put less emphasis on organisational structure but focus on the dynamics in the sense of implementing and experimenting with simulation models. Often these simulation models are based on some implementation environment and not specified in an implementation-independent manner using a formally defined conceptual language. The Strictly Declarative Modelling Language SDML [38] (see also [9]) , and the use of the agent-oriented modelling approach DESIRE in social simulation as presented in [8] are some of the few exceptions. Both modelling approaches focus on specification and simulation; however, they do not offer dedicated support for a specific type of organisational structure. Moreover, simulation of dynamics is the main purpose; not much formally defined support is offered for analysis of dynamics, such as checking whether a given simulation or empirical trace satisfies a given dynamic property.

Thirdly, temporal modelling is one of the dominant approaches for specification and analysis of dynamic properties in agent systems in general; e.g., [1], [9], [15], [16], [17], [28], [33], [35], [38]; see also [18], [19] for an overview. One of the strong points in this area of research is the declarative modelling of simulation models, for example based on the paradigm of Executable Temporal Logic [1]. Also formal analysis (verification) can be supported using this type of formalisation; e.g., [9], [16], [33], [35]. However, the temporal languages or logics usually adopted do not have means to explicitly reflect organisational structure in an organisational behaviour specification.

The Agent/Group/Role (AGR) approach (previously called Aalaadin) introduced in [12] is an example of an approach initially focussing on organisational structure, abstracting from the details of the dynamics. However, [13] and [14] are some first steps to relate specifications of dynamic properties to the organisational structure provided by AGR. In [1] the MOCA system presented combines the AGR model with Madkit into a platform, based on a theoretical foundation. It is shown how dynamics of the organisational structure itself can be modelled: agents that can dynamically create, join, or quit groups. As this AGR approach will be used in more detail in later sections, it will be introduced in some more detail

## 2.2 The AGR Organisation Modelling Approach

To model an organisation, the Agent/Group/Role (AGR) approach [12] can be used. Within this organisation modelling approach an organisation structure is described at three aggregation levels: the *organisation* consists of a set of *groups*, and each group consists of the *roles* in that group. Furthermore, *connections* between roles and between groups are possible. Moreover the organisation is *realized* by *agents* fulfilling roles (not depicted). The main concepts are briefly described as follows:
- The *agents*. The model places no constraints on the internal architecture of agents. An agent is only specified as an active communicating entity which plays roles within groups. This agent definition is intentionally general to allow agent designers to adopt the most accurate definition of agent-hood relative to their application.

- A *group* is defined as an atomic set of agent roles. Each agent takes part by fulfilling roles in one or more groups. In its most basic form, the group is only a way to name a set of roles.
- A *role* is an abstract representation of an agent function or service within a group. Each agent can handle multiple roles, and each role handled by an agent is local to a group.

The *organisational structure* is the specification of a specific multi-agent organisation based on a definition of groups, roles and their relationships within the organisation. A central notion is the *group structure*. It identifies the roles and (intragroup) interaction between roles within a group. The group structure is defined by the set of roles, interaction schemes and an (intragroup) transfer or communication model within the group. In addition, intergoup role relations between roles of different groups specify the connectivity of groups within an organisation.

As an illustration, an example of a bank service provision organisation using a Call Centre is used. From an organisation modelling perspective, an organisation model can be defined using the following two groups: Client Service (sometimes also called Open Group) and Distribution. The roles within the groups are as follows:

*Client Service*: Receptionist, Client
*Distribution*: Distributor, Participants

Within the Client Service group the service requests of clients are identified in an interaction between Client Relator and Client (this takes place within the Call Centre). See Figure 1; in this figure

- the big ovals denote *group instances*,
- small grey ovals denote *role instances* within a group instance,
- arrows denote *interactions between role instances* in *one* group instance, and
- dashed vertical lines denote *relations* between role instances in *different* group instances.

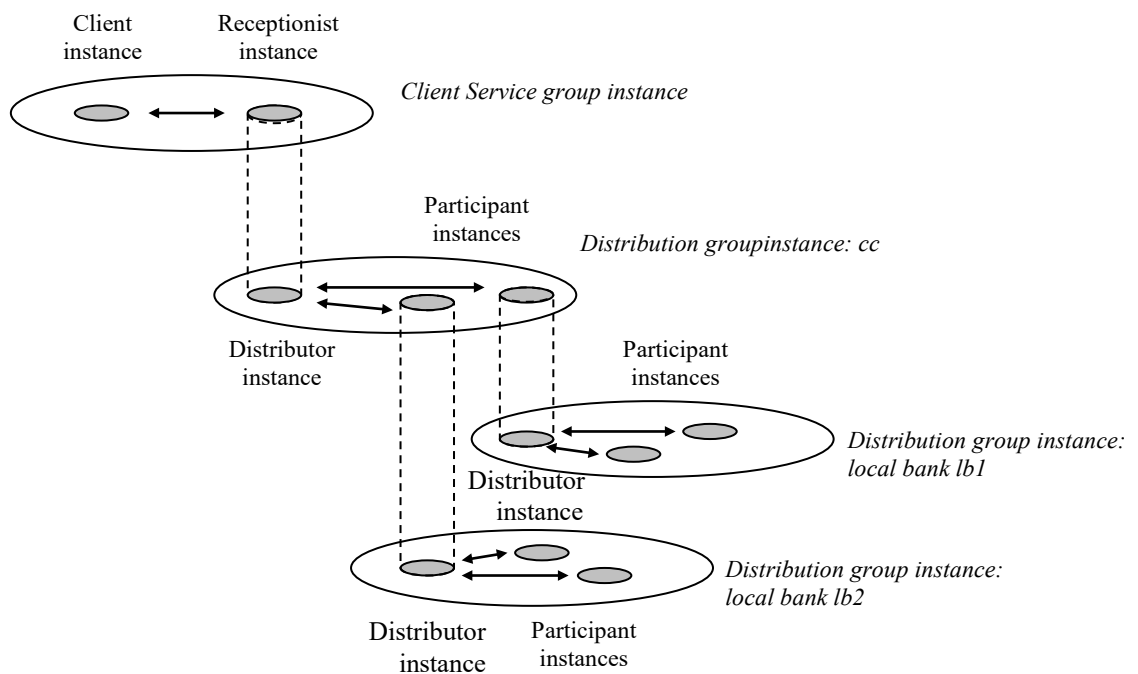

**Fig. 1.** The AGR-organisation structure for a bank Call Center organisation

Within a Distribution group, an identified service request is allocated in mutual interaction between Distributor and Participants. Actually this process takes place in two steps, making use of different instantiations of the Distribution group: once in a first line Distribution group (relating a Call Centre agent to local bank work manager agents) and once in second line Distribution groups (work manager and employees within a local bank). The agents with role Participant in the first line Distribution group have the role of Distributor in a second line Distribution group. Also an agent with role Receptionist in the Client Service group has a role of Distributor in the first line Distribution group. At the level of the organisation model, this means that between these pairs of roles within the organisation model, intergroup role relations are defined.

## 3 Formal Specification of Organisation Structure

This section presents an approach to a foundation for the specification of organisation *structure*. Organisation structure is often depicted in diagrammatic form (for example, as kind of labelled graph; e.g., see Figure 1 or 2) consisting of different aggregation levels and different types of elements within the organisation (such as roles, groups, interactions), and relationships between these elements. A suitable formalisation approach for such structure descriptions is the notion of semantic structures (or models, in terms of logic) for many-sorted predicate logic; e.g., [36]. These structures will be denoted by tuples

$$< S_1, \ldots S_n; R_1, \ldots R_p; F_1, \ldots, F_q >$$

where $S_i$ are sorts, $R_j$ relations over sorts, and $F_k$ functions on sorts. This formalisation approach is adopted as a foundation for specification of organisational structure, and illustrated for AGR organisation structures.

As an example a factory is considered that is organised at the highest aggregation level according to two divisions: *division A* that produces certain components and *division B* that assembles these components to (composite) products. At one aggregation level lower the division A is organised according to two departments: *department A1* (the work planning department for division A) and *department A2* (component production department). Similarly, division B is organised according to two department roles: *department B1* (for assembly work planning) and *department B2* (product production department).

The two divisions are modeled as *groups* (depicted by the larger ovals), with the departments as their *roles* (depicted by smaller ovals within larger ones). A third group, the Connection Group C, models the communication between the two divisions. This group consists of the two *roles* 'division A representative' and 'division B representative'. *Intergroup role interactions* (depicted by pairs of dotted lines) are modeled between the role 'department A1' in the division A group and the role 'division A representative' within the connection group, and between the role 'department B1' in the division B group and the role 'division B representative' within the connection group. *Intragroup role transfers* model communication between the two roles within each of the groups (depicted by the arrows).

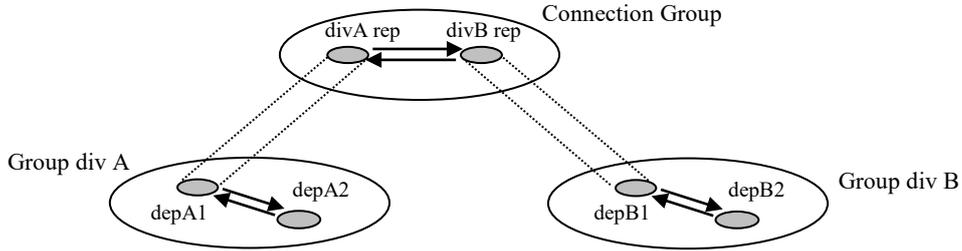

**Fig. 2.** Example AGR organisation structure used to illustrate formalisation

Connections have destination roles (indicated by the arrow points) and source roles (no pointing). Based on the semantic structures of many-sorted predicate logic a more precise formal definition of an AGR structure is the following.

**Definition 1  (AGR Organisation Structure)**
Let IDENT be the set of all identifiers, and let Groups, Roles, Transfers, and Intergroup_interactions be subsets of IDENT which subsequently contain the names of groups, roles, transfers, and intergroup interactions in the organisation. An AGR *organisation structure* is defined by a tuple of sets and relations in the following manner:

    AGROrg  =  < Groups, Roles, Intergroup_interactions, Transfers,
                  role_in, source_of_interaction, destination_of_interaction,
                  source_of_transfer, destination_of_transfer >

Each group involves a set of roles. The relationship

    role_in:        Roles × Groups

defines which role is in which group. The source and destination roles that an intergroup interaction connects are specified by

    source_of_interaction, destination_of_interaction:  Roles × Intergroup_interactions

Furthermore, transfers relate source roles to destination roles. The relationships

    source_of_transfer, destination_of_transfer:    Roles × Transfers

specify the roles a transfer connects. It is demanded that source and destination of a transfer belong to the same group:

∀sr, dr ∈ Roles, ∀ t ∈ Transfers:

  ( source_of_transfer(sr, t) ∧ destination_of_transfer(dr, t) ) ⇒ ∃g ∈ Groups: ( role_in(sr, g) ∧ role_in(dr, g) )

The example organisation of Figure 1 has
    Groups               = {divA, divB, C},
    Roles                = {depA1, depA2, depB1, depB2, divArep, divBrep},
    Intergroup_interactions  = {iAC, iCA, iBC, iCB}
    Transfers          = {tA12, tA21, tB12, tB21},

An overview of the relationships depicted in Figure 1 is:

| *within divA , divB and C* | *organisation level* |
|---|---|
| role_in(depA1, divA)<br>role_in(depA2, divA) | source_of_interaction(depA1, iAC)<br>destination_of_interaction(divArep, iAC) |
| role_in(depB1, divB) | source_of_interaction(divArep, iCA) |
| role_in(depB2, divB) | destination_of_interaction(depA1, iCA) |
| role_in(divArep, C) | |
| role_in(divBrep, C) | source_of_interaction(depB1, iBC) |
| | destination_of_interaction(divBrep, iBC) |
| source_of_transfer(depA1, tA12) | |
| destination_of_transfer(depA2, tA12) | source_of_interaction(divBrep, iCB) |
| | destination_of_interaction(depB1, iCB) |
| source_of_transfer(depA2, tA21) | |
| destination_of_transfer(depA1, tA21) | |
| source_of_transfer(depB1, tB12) | |
| destination_of_transfer(depB2, tB12) | |
| source_of_transfer(depB2, tB21) | |
| destination_of_transfer(depB1, tB21) | |
| source_of_transfer(divArep, tC12) | |
| destination_of_transfer(divBrep, tC12) | |
| source_of_transfer(divBrep, tC21) | |
| destination_of_transfer(divArep, tC21) | |

Note that, for simplicity, no difference is made between a role and role instances that inherit properties of the role. When desired, within a specific language developed on the basis of the formalisation presented here such a difference can be made. Furthermore, intergroup interactions are defined between two roles; this can easily be generalised for intergroup interactions involving more than two roles.

For all i ∈ Intergroup_interactions (resp. t ∈ Transfers or g ∈ Groups), let involved_roles(i) (resp. involved_roles(t), and involved_roles(g)) denote the set of all roles that are involved in interaction i (resp. transfer t or group g):

involved_roles(i) = { r ∈ Roles | destination_of_interaction(r, i) ∨ source_of_interaction(r, i) }
involved_roles(t) = { r ∈ Roles | destination_of_transfer(r, t) ∨ source_of_transfer(r, t) }
involved_roles(g) = { r ∈ Roles | role_in(r, g) }

## 4 Formal Specification of Organisation Behaviour Based on Organisation Structure

After a foundation of an organisation structure has been defined, foundations for specification of dynamic properties in an organisation are addressed. The aim is not only to cover simple types of dynamics, such as simple reactive behaviour, but also more complex dynamics. For specification of more dynamic properties, often temporal logical languages are used; such language have no internal structuring (other than the manner in which formulae can be formed by logical connectives).

### 4.1 From Organisation Structure to Sets of Dynamic Properties for Organisation Behaviour

The challenge here is to incorporate somehow the organisational structure within the logical description of the organisation's internal dynamics. Just specifying one logical theory as a set of dynamic (temporal) properties describing the behaviour of the organisation would not be fully satisfactory, as in that case the organisational structure is lost in the dynamics specification or at least remains implicit. To capture the organisation structure within the organisation behaviour description, at least the different aggregation levels and more in general the different elements of the or-ganisation structure are to be addressed in an explicit manner. To this end the following approach is introduced:

- for each *element* within the organisational structure characterise its dynamics by a *specific set of dynamic properties*; this is addressed in Section 4
- based on *structural relations between elements* in an organisational structure, identify *relationships between the sets of dynamic properties* corresponding with these elements; this is addressed in Section 5

In general, the dynamics of an element within an organisation structure can be characterised by specification of dynamic properties expressing relationships of states of that element over time. For a role the concept 'state' needs to be defined both for the input and the output of the role. Since transfers and intergroup interactions are assumed to operate only on input and output states of roles, without having their own internal state, no further state is assumed for transfers and intergroup interactions. To define states the notion of state property is useful, which is expressed in terms of a state ontology. Moreover, the notion of trace as a sequence of states over a time frame is used to formalise dynamics.

**Definition 2 (Ontology, State, Trace)**
(a) A *state ontology* is a specification (in order-sorted logic) of a vocabulary, i.e., a signature. A state for ontology Ont is an assignment of truth-values {true, false} to the set At(Ont) of ground atoms expressed in terms of Ont. The *set of all possible states* for state ontology Ont is denoted by STATES(Ont).
(b) A fixed *time frame* T is assumed which is linearly ordered. A *trace* $\mathcal{T}$ over a state ontology Ont and time frame T is a mapping $\mathcal{T}$: T → STATES(Ont), i.e., a sequence of states $\mathcal{T}_t$ (t ∈ T) in STATES(Ont). The set of all traces over state ontology Ont is denoted by TRACES(Ont).

Depending on the application, it may be dense (e.g., the real numbers), or discrete (e.g., the set of integers or natural numbers or a finite initial segment of the natural numbers), or any other form, as long as it has a linear ordering.

**Definition 3 (State Properties and Dynamic Properties)**
Let Σ be a given set of state ontologies.
(a) The set of *state properties* STATPROP(Σ) is the set of all propositions over ground atoms expressed in the ontologies from Σ.
(b) Let L be a language for dynamic properties. The set of *dynamic properties* DYNPROP$_L$(Σ) is the set of formulae that can be formulated in language L with respect to traces based on the set of state ontologies Σ.

The subscript L is dropped, when no confusion is expected and when the usage of the set is irrespective of a choice of language. For the paper such a language L is assumed with semantic consequence relation ⊨. The approach is independent of the choice of this language. In the next subsection two examples of languages that can be used are discussed.

**4.2 Two Example Languages for Dynamic Properties**

From a philosophical perspective [18] considers two main streams in temporal logic: modal logic approaches to temporal logic (developed mainly within Computer Science; e.g., [9], [15], [16], [17]), and predicate logic approaches to temporal logic (developed mainly within AI). In [19] different approaches in the latter stream are addressed in more detail. Two substreams distinguished are the use of additional temporal arguments within domain predicates, and the reification ap-proach, where state properties are represented not by statements but by terms in the language, and predicates are used to express temporal structure over these term expressions; examples are event calculus [30], situation calculus [40] and the language TTL discussed below. In this approach part of the model theory is incorporated in the language. This reification approach to predicate logical temporal modelling is the approach adopted here. Two examples of languages that can be used to express dynamic properties that are considered in some more detail are:

- LTL (Linear Time Logic)
  This language is one from the stream based on modal logic approaches (cf. [15], [17], [9]); to make it suitable it is extended with operators that are specific for elements of the organisational structure or multi-agent system
- TTL (Temporal Trace Language)
  This language is one from the stream based on predicate logic approaches, and in particular the substream of reification approaches; e.g., [28], [16]

With in TTL, explicit reference is made to the part of the organisation within the state-function. Therefore, for TTL the set is defined relative to a set of identifiers to parts of the organisation structure and a set of ontologies. The set of identifiers can, for example, be the set of role names that can occur in the dynamic properties:

**Definition 4   (DYNPROP$_{TTL}$)**
The set DYNPROP$_{TTL}$(R, $\Sigma$) is defined inductively as follows:
- For all r $\in$ R, for all parts P within O containing r, all traces $\gamma \in$ Traces(ONT(P)), for all t $\in$ T, and for all state properties s $\in$ STATPROP($\Sigma$), the formulae (here $\models$ is an infix predicate symbol):
    state($\gamma$, t, r) $\models$ s
    state($\gamma$, t, input(r)) $\models$ s
    state($\gamma$, t, output(r)) $\models$ s
  are elements of DYNPROP$_{TTL}$(R, $\Sigma$).
- If $\varphi \in$ DYNPROP$_{TTL}$(R, $\Sigma$) and $\psi \in$ DYNPROP$_{TTL}$(R, $\Sigma$), then also
    $\varphi \wedge \psi$, $\varphi \vee \psi$, $\neg\varphi$, $\varphi \Rightarrow \psi \in$ DYNPROP$_{TTL}$(R, $\Sigma$).
- Quantification is allowed in the normal way (e.g., over time, traces, parts and state properties).
- Mathematical operations and equations can be used in the normal way.

Typically, in terms of TTL, each role dynamic property $\varphi$ of role r contains at least one reference to both the terms input(r) and output(r). For a linear temporal logic, this entails that the role name is used as index for some modal operator within the property. Furthermore, the properties only refer to r and to concepts from the ontology used in the interfaces of r.

Next, consider the linear time temporal logic LTL with the usual modal operators H (always in the past), P (at some time in the past), C (currently), X (next), F (at some time in the future), G (always in the future). The modal operators can be extended with a temporal parameters constraining the operator; e.g., $F^{<10}$(prop) means that property prop will hold somewhere in the future before 10 time units have passed. The expression $F^{10}$(prop) means that property prop will hold exactly after 10 time units. The modal operators can be indexed with either a group or a role name to indicate which part of the information state is meant. For example, $C_r$(finished) means that currently in the information state of the role r the proposition answer holds. For a formalization of compositional multi-agent systems using temporal multi-epistemic logic, see (Engelfriet, Jonker & Treur, 2002).

**Definition 5   (DYNPROP$_{LTL}$)**
The set DYNPROP$_{LTL}$(R, $\Sigma$) is defined inductively as follows:

- For all r ∈ R, for all time constraints tc, for all propositions s ∈ Σ, for all modal operators M, the formula $M_r^{tc}(s)$ is an element of DYNPROP$_{LTL}$(R, Σ).
- If φ ∈ DYNPROP$_{LTL}$(R, Σ) and ψ ∈ DYNPROP$_{LTL}$(R, Σ), then also φ ∧ ψ, φ ∨ ψ, ¬φ. φ ⇒ ψ ∈ DYNPROP$_{LTL}$(R, Σ).

In Table 1 some simple examples of dynamic properties in the two languages are shown.

| Informal Dynamic Properties | Formal Dynamic Properties in DYNPROP$_{TTL}$ and DYNPROP$_{LTL}$ |
|---|---|
| *Role property of* r1:<br><br>Requests received by r1 are answered by r1 within 10 seconds. | In TTL:<br>∀ t [ state(γ, t, output(r1)) ⊨ request<br>    ⇒ ∃t': t ≤t'≤t +10 ∧state(γ, t', output(r1)) ⊨ answer ]<br>In LTL:<br>$C_{r1}$(request) ⇒ $F_{cl2}^{\leq 10}$(answer) |
| *Transfer property for transfer between* r1 *and* r2:<br><br>Information available on the output of r1 arrives at the input of r2 within 1 second. | In TTL:<br>∀ t [ state(γ, t, output(r1)) ⊨ answer<br>    ⇒ ∃ t': t ≤ t'≤t+1∧ state(γ, t', input(r2)) ⊨ answer ]<br>In LTL:<br>$C_{r1}$(answer) ⇒ $X_{r2}$(answer) |
| *Intergroup role interaction* I1:<br><br>Requests received at its input by r1 in group g1 will become available within group g2 on the output of role r2 within 1 second. | In TTL:<br>∀ t [ state(γ, t, input(r1)) ⊨ request<br>    ⇒ ∃ t': t ≤ t'≤t+1∧ state(γ, t', output(r2)) ⊨ request ]<br>In LTL:<br>$C_{r1}$(request) ⇒ $X_{r2}$(request) |
| *Relative Adaptivity*:<br><br>'Experience leads to quality': The more intensively requests have been handled by r, the higher the quality. | In TTL:<br>[∀w1, w2 state(γ1, 0, output(r)) ⊨ has_level(quality, w1) &<br>state(γ2, 0, output(r)) ⊨ has_level(quality, w2)  ⇒ w1≤ w2 ] &<br>[ ∀t [ state(γ1, t, input(r)) ⊨ has_intensity(requests, v1) &<br> state(γ2, t, input(r)) ⊨ has_intensity(requests, v2)    ⇒ v1≤ v2 ]<br>⇒ ∀t [∀w1, w2 state(γ1, t, output(r)) ⊨ has_level(quality, w1) &<br>    state(γ2, t, output(r)) ⊨ has_level(quality, w2) ⇒ w1≤ w2 ] ]<br>It is not possible to express this in LTL. |

**Table 2. Example Dynamic Properties in TTL and LTL**

### 4.3 Organisation Dynamics for an AGR Organisation Structure

In order to characterise the dynamics within an organisation, dynamic properties have to be specified for each of the aggregation levels and, more specifically, for the elements of an organisation structure: for the AGR modelling approach for each role, each transfer, each group, each intergroup interaction, and for the organisation as a whole. A specification of the dynamics requires a specification of the state ontologies used (for expressing state properties) for input states and out-put states of roles. The specifications of the dynamic properties are based on the given state ontologies.

**Definition 6   (AGR Organisation Dynamics)**

Let ONT be a set of (state) ontologies and O an AGR organisation structure defined by

    O   =   < Groups, Roles, Transfers, Interactions,
                  role_in, source_of_transfer, destination_of_transfer,
                  source_of_interaction, destination_of_interaction >.

The dynamics of the AGR organisation O over ONT is formalised by a tuple as follows:

AGRDyn = < O, role_input_ontologies, role_output_ontologies,
          role_dynproperties, transfer_dynproperties, interaction_dynproperties >

where

    role_input_ontologies, role_output_ontologies:      Roles    → ℘(ONT)

        ontologies for input resp. output of each role r

    role_dynproperties:                            Roles    → ℘(DYNPROP(ONT))

        dynamic properties for each role r

    transfer_dynproperties:                     Transfers → ℘(DYNPROP(ONT))

        dynamic properties for each transfer t

    group_dynproperties(g)                   Groups → ℘( DYNPROP(ONT))

        dynamic properties for each group g

    intergroup_interaction_dynproperties:      Intergroup_Interactions → ℘(DYNPROP(ONT))

        dynamic properties for each intergroup interaction i.

    organisation_dynproperties:                 {O} → ℘( DYNPROP(ONT))

        dynamic properties for the organisation O.

Thus the dynamics of the AGR organisation O is specified by: ontology sets for each role r, sets of dynamic properties for each transfer t, group g and intergroup interaction I, and the organisation as a whole. For these mappings the constraints C1 … C5 listed below are assumed to be fulfilled.

**C1  Role dynamic properties**
Role dynamic properties relate input to output of that role:

    ∀r ∈ Roles: role_dynproperties(r) ⊆ DYNPROP(r, ONT(r))

For example, the gossip role behaviour: 'whenever somebody tells you something, you will tell it to everybody else' is expressed in terms of input of the role leading to output of the role in a reactive manner. An example relating to Figure 1:

**DP(depA1)        Progress Information Generates Planning in depA1**
  If within division A department A1 receives progress information on component production,
  then an updated planning will be generated by department A1 taking this most recent information into account.

**C2  Transfer dynamic properties**
Transfer properties relate output of the source roles to input of the destination roles:
∀ t ∈ Transfers: transfer_dynproperties(t) ⊆

DYNPROP( involved_roles(t) , ∪{ role_output_ontologies(r) | source_of_transfer(r, t) }
∪ ∪{ role_input_ontologies(r) | destination_of_transfer(r, t) }   )

Typically, such sets contain properties like, information is indeed transferred from source to destination, transfer is brought about within x time, arrival comes later than departure, and information departs before other information also arrives before that other information.

### C3  Group dynamic properties
Group dynamic properties relate input and/or output of roles within a group.

group_dynproperties(G) ⊆ DYNPROP(G, ONT(G))

An example of a group property is: "if the manager asks anyone within the group to provide the secretary with information, then the secretary will receive this information". A special case of a group property is an *intragroup interaction* relating the outputs of two roles within a group. A typical (informal) example of such an intragroup interaction property is: "if the manager says 'good afternoon', then the secretary will reply with 'good afternoon' as well". Other examples may involve statistical information, such as "3 out of the 4 employees within the organisation never miss a committed deadline". Example relating to Figure 1 for division A:

**DP(A)   A Progress Information Generation**
  This property is the conjunction of the following properties.

**DP1(A)   Initial A Progress Information Generation**
  Department A1 receives initial progress information on component production processes, involving already available components.

**DP2(A)   Subsequent A Progress Information Generation**
  Within the division A group, for any component production planning generated by department A1, incorporating a specific required set of components, progress information on the production of these components will be received by department A1.

### C4  Intergroup interaction dynamic properties
Intergroup interaction properties relate the input of the source role to the output of the destination role:

∀ i ∈ Intergroup_interactions:
  intergroup_interaction_dynproperties(i) ⊆ DYNPROP( involved_roles(i) ,
    ∪{ role_input_ontologies(r) | source_of_interaction(r, i)  }
    ∪ ∪{ role_output_ontologies(r) | destination_of_interaction(r, i) } )

For example, a project leader is asked by one of the project team members (input of role 'project leader' within the project group) to put forward a proposal in the meeting of project leaders (output of role 'member' within the project leaders group). An example relating to Figure 1:

**Intergroup Role Interaction between A and C: IrRI(A, C)**
For the connectivity between the groups A and C, the following intergroup role interaction properties are considered, one from A to C, and one from C to A.

**IrRI(depA1, divArep)        Progress Information Provision A to B**
  If within division A progress information on component production is received by department A1, then within the connection group this will be communicated by the division A representative to the division B representative.

**IrRI(divArep, depA1)        B Progress Information Incorporation by A**

If within the connection group the division A representative receives information from the division B representative on which components are needed, then within division A a component production planning will be generated by department A1 taking these into account.

**C5  Organisation dynamic properties**
Organisation dynamic properties relate to input and/or output of roles within the organisation.

organisation_dynproperties(G) ⊆ DYNPROP(O, ONT(O))

A typical (informal) example of such a property is: "if within the organisation, role A promises to deliver a product, then role B will deliver this product". An example relating to Figure 1:

**DP(F)   Overall Progress Notification**
If a request for a product is made (by a client),
then progress information will be provided (for the client).

The different types of dynamic properties all relate to different combinations of input and output. Table 1 provides an overview of these combinations. Note that with respect to simulation, the above dynamics definition can contain elements that are redundant: a smaller subset of dynamical properties can form an executable specification of the dynamics of an AGR type organisation. For example, on the basis of the role and transfer dynamic properties and intergroup interactions the organisation can be simulated. The group dynamic properties, including the intragroup role interaction properties, and the organisation properties should emerge in the execution, and testing for them can validate the model.

**Table 1.** Types of Dynamic Properties for an AGR Organisation Model

| *Dynamic Property Type* | Relating | |
|---|---|---|
| **Role r** | Role r Input | Role r Output |
| **Transfer from r1 to r2** | Role r1 Output | Role r2 Input |
| **Group G** | Input or Output of roles in G | |
| **Intragroup interaction** | Role r1 Output | Role r2 Output |
| **Intergroup interaction** | Role r1 Input | Role r2 Output |
| **Organisation** | Input or Output of roles in O | |

In order to make an executable organisation model the dynamical properties need to be chosen from the set of executable dynamical properties EXEDYNPROP ⊆ DYNPROP, for example Executable Temporal Logic [1], or the 'leads to' format presented in [29], [2].

## 5  Interlevel Relations between Dynamics at Different Aggregation Levels

An organisational structure defines relations between different elements in an organisation. In Section 4 the dynamics of these different elements were characterised by sets of dynamic properties. An organisational structure has the aim to keep the overall dynamics of the organisation manageable; therefore the structural relations between the different elements within the organisational structure have to impose somehow relationships or dependencies between their dynamics. In the introduction to their book Lomi and Larsen [32] emphasize the importance of such relationships. Organisations can be seen as adaptive complex information processing systems of (boundedly) rational agents, and as tools for control; central challenges are [32]:

- from the first view: 'given a set of assumptions about (different forms of) individual behaviour, how can the aggregate properties of a system be determined (or predicted) that are generated by the repeated interaction among those individual units?'
- from the second view: 'given observable regularities in the behaviour of a composite system, which rules and procedures - if adopted by the individual units - induce and sustain these regularities?'.

Both views and problems require means to express relationships between dynamics of different elements and different levels of aggregation within an organisation. In [32] two levels are mentioned: the level of the organisation as a whole versus the level of the units. Also in the development of MOISE (cf. [16], [23], [26]) an emphasis is put on relating dynamics to structure. Within MOISE dynamics is described at the level of units by the goals, actions, plans and resources allocated to roles to obtain the organisation's task as a whole. Specification of the task as a whole may involve achieving a final (goal) state, or an ongoing process (maintenance goals) and an associated plan specification.

As in our formalisation introduced in Section 4, dynamics are characterised by sets of dynamic properties for the respective elements of the organisation, the next step to be made is to identify how organisational structure determines (mathematically defined) relationships between these sets of dynamic properties for the different elements and aggregation levels within an organisation. Preferably such relations between sets of dynamic properties would be of a logical nature; this would allow the use of logical methods to analyse, verify and validate organisation dynamics in relation to organisation structure. Indeed, in our approach presented below, logical relationships between sets of dynamic properties of elements in an organisation turn out an adequate manner to (mathematically) express such dynamic cross-element or cross-level relationships. This will be illustrated for the AGR approach. Within AGR organisation models three aggregation levels are involved:
- the organisation as a whole; the highest aggregation level.
- the level of a group
- the level of a role within a group

A general pattern for the dynamics in the organisation as a whole in relation to the dynamics in groups is as follows:

> dynamic properties for the groups &
> dynamic properties for intergroup role interaction
> $\Rightarrow$ dynamic properties for the organisation

Moreover, dynamic properties of groups can be related to dynamic properties of roles as follows:

> dynamic properties for roles &
> dynamic properties for transfer between roles
> $\Rightarrow$ dynamic properties for a group

The idea is that these are properties dynamically relating a number of roles within one group. To get the idea, consider the special case of an intragroup role interaction from role r1 to role r2, characterised by dynamic properties, that relate output of one role r1 to output of another role r2. Assuming that transfer from output of r1 to input of r2 is adequate and simply copies the information, this property mainly depends on the dynamics of the role r2.

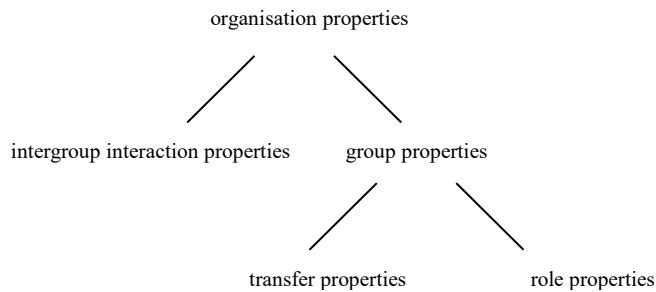

**Fig. 3.** Overview of interlevel relations between dynamic properties

Therefore in this case the relationship has the form:
  dynamic properties for role r2 &
  dynamic properties for transfer from role r1 to role r2
  $\Rightarrow$ dynamic properties of intragroup interaction from r1 to r2

An overview of the logical relationships between dynamic properties at different aggregation levels is depicted as an AND-tree in Figure 3. The logical relationships put forward above can be formalised in the following manner. In Definition 5 below the following notations are used, where G is a group:

 con(F)       is the conjunction of all dynamic properties in a finite set F
 role_dynproperties(G)  = $\cup$ { role_dynproperties(t) | role_in(r, G) }
 transfer_dynproperties(G)  =
  $\cup$ { transfer_dynproperties(t) | t: Transfer, involved_roles(t) $\subseteq$ involved_roles(G) }
 transfer_dynproperties(O)  = $\cup$ { transfer_dynproperties(t) | t: Transfer }
 intergroup_interaction_dynproperties(O) =
  $\cup$ { intergroup_interaction_dynproperties(i) | i: Intergroup_interaction }

**Definition 7 (Role-Group and Group-Organisation Interlevel Relations)**
Let AGRDyn be a model for the dynamics of an AGR organisation structure O.
(a) A *logical role-group interlevel relation* for group G is a logical statement of the form
    $DP_1$ & ... & $DP_n$ & TR   $\Rightarrow$  DP
 where
  DP  a group dynamics property for G, from     group_dynproperties(G)
  $DP_i$  a role dynamics property or conjunction thereof from  role_dynproperties($r_i$)
  TR  a transfer dynamics property or conjunction thereof from  transfer_dynproperties(G)

The set of all role-group interlevel relations for G is denoted by RGIR(G); the union of all of them over all groups is denoted by RGIR.

(b) A *logical group-organisation interlevel relation* for O is a logical statement in of the form
    $DP_1$ & ... & $DP_n$ & TR & IID   $\Rightarrow$  DP
 where
  DP  an organisation dynamics property from     org_dynproperties(O)
  $DP_i$  a group dynamics property or conjunction thereof from
    group_dynproperties($G_i$)
  TR  a transfer dynamics property or conjunction thereof from
    transfer_dynproperties(O),
  IID  an intergroup interaction property or conjunction thereof
    from intergroup_interaction_dynproperties(O)

 The set of all group-organisation interlevel relations is denoted by GOIR.

(c) A *logical interlevel relation assignment* for an AGR organisation model AGRDyn is a mapping
    interlevel_relations:     {O} $\cup$ Groups $\rightarrow$ $\wp$(RGIR $\cup$ GOIR)

such that the set interlevel_relations(O) consists of logical group-organisation interlevel relations for O (i.e., from GOIR), and for each group G the set interlevel_relations(G) consists of logical role-group interlevel relations for G (i.e., from RGIR(G)).

(d) The *standard logical interlevel relation assignment*
    standard_interlevel_relations:   {O} $\cup$ Groups $\rightarrow$ $\wp$(RGIR $\cup$ GOIR)
for an AGR organisation model AGRDyn is the mapping defined by:
standard_interlevel_relations(G) is the set of logical relations

{ con(role_dynproperties(G)) ∪ transfer_dynproperties(G)) ⇒ DP | DP∈group_dynproperties(G) }
standard_interlevel_relations(O) is the set of logical relations
{ con(role_dynproperties(O) ∪ transfer_dynproperties(O) ∪
    intergroup_interaction_dynproperties(O)) ⇒ DP | DP ∈ organisation_dynproperties(O) }.

Notice that this definition provides a formalisation for the type of relations that Lomi and Larsen [32] put forward as one of the challenges in organisation modelling (see above). Following the general implication pattern given above, the following specific relationships between dynamic properties at different aggregation levels can be established:

*From group properties and intergroup role interaction properties to organisation properties*

    DP(A) & DP(B) & DP(C) & IrRI(F)        ⇒        DP(F)

*From intragroup role interaction properties to group properties*

    IaRI(A) & TRD(A)        ⇒        DP(A)
    IaRI(B) & TRD(B)        ⇒        DP(B)
    TRD(C)                  ⇒        DP(C)

*From dynamic role properties to intragroup interaction properties*

    DP(depA1) & DP(depA2) & TRD(A)        ⇒        IaRI(A)
    DP(depB1) & DP(depB2) & TRD(B)        ⇒        IaRI(B)

Such relationships between dynamic properties can be visualised in the form of an AND-tree; see Figure 4 (the names have been kept short to keep the picture concise).

Given these definitions notions such as 'valid' and 'complete' can be defined and a proposition can be formulated.

**Definition 8 (Valid and Complete)**
Let AGRDyn be a model for the dynamics of an AGR organisation structure O
(a) A logical interlevel relations assignment interlevel_relations is called *valid* if all involved logical interlevel relations are valid statements in the logic used, i.e., for any trace, if the antecedent holds, also the consequent holds.
(b) The model AGRDyn is called *grounded* if its standard logical interlevel relation assignment is valid.
(c) A logical interlevel relations assignment interlevel_relations is called *connected* if for any group G each group property occurring (in an antecedent) in interlevel_relations(O) also occurs (as a consequent) in interlevel_relations(G).
(d) A logical interlevel relations assignment interlevel_relations is called *complete* if for any group G each group property from group_dynproperties(G) occurs (as a consequent) in interlevel_relations(G) and each organisation property from organisation_dynproperties(O) occurs (as a consequent) in interlevel_relations(O).

Notice that the standard logical interlevel relation assignment is connected and complete. However, the antecedents used are not minimal in the sense that many irrelevant conjuncts may occur in them.

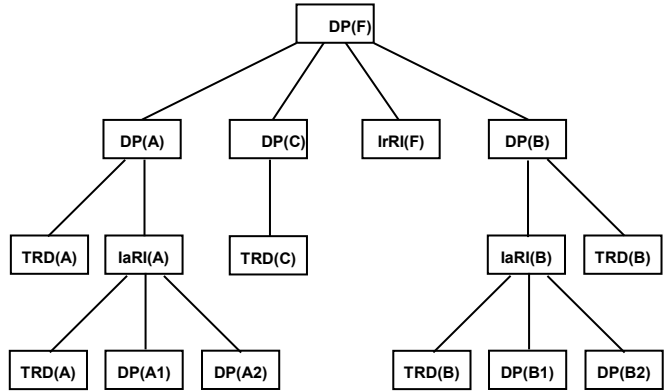

**Fig. 4.** Example interlevel relations between dynamic properties in the form of an AND-tree

**Proposition**
Let AGRDyn be a model for the dynamics of an AGR organisation structure O, and interlevel_relations an interlevel relation assignment. Moreover, let $\mathcal{T}$ be a trace that satisfies all dynamic properties in role_dynproperties(r) for all roles r, all transfer properties in transfer_dynproperties(t) for all transfers t, and all intergroup interaction properties in intergroup_interaction_dynproperties(i) for all intergroup interactions i.
(a) If the interlevel relation assignment interlevel_relations is valid, then trace $\mathcal{T}$ satisfies all dynamic group properties occurring in interlevel_relations(G) for any group G.
(b) If the interlevel relation assignment interlevel_relations is valid and complete, then the trace $\mathcal{T}$ satisfies all dynamic group and organisation properties in organisation_dynproperties(O) and group_dynproperties(G) for any group G.
(c) If a valid and complete logical interlevel relations assignment for AGRDyn exists, then AGRDyn is grounded.

This Proposition implies the following for an organisation model with a valid and complete interlevel relation assignment. If the properties for roles, transfers and intergroup interactions are in executable format, and used for simulation (e.g., based on the paradigm of Executable Temporal Logic), then a generated trace will satisfy these properties, and, hence by the Proposition satisfy all group and organisation properties as well. Among others, this gives means to validate (in the sense of falsification) an organisation model.

## 6 Organisation Realisation

In this section criteria are discussed when allocation of a set of agents to roles is appropriate to realize the organisation dynamics, illustrated for the AGR approach. One of the advantages of an organisation model is that it abstracts from the specific agents fulfilling the roles. This means that all dynamic properties of the organisation remain the same, independent of the particular allocated agents. However, the behaviours of these agents have to fulfil the dynamic properties of the roles and their interactions. The organisation model can be (re)used for any allocation of agents to roles for which:
- for each role, the allocated agent's behavior satisfies the dynamic role properties,

- for each intergroup role interaction, one agent is allocated to both roles and its behavior satisfies the intergroup role interaction properties, and
- the communication between agents satisfies the respective transfer properties.

Expressed differently, for a given allocation of agents to roles the following logical relationships between dynamic properties hold:

*agent – role*
from dynamic agent properties to dynamic role properties:

>   agent A is allocated to role r &
>   dynamic properties of agent A          $\Rightarrow$
>           dynamic properties of role r

*agent – intergroup role interaction*
from dynamic agent properties to dynamic intergroup role interaction properties:

>   agent A is allocated to roles r1 and r2 in different groups &
>   dynamic properties of agent A          $\Rightarrow$
>           dynamic properties of intergroup role interaction between r1 and r2

*agent communication – role transfer*
from dynamic agent communication properties to dynamic transfer properties:

>   agent A is allocated to role r1 and agent B to role r2 in one group &
>   dynamic properties of communication from A to B $\Rightarrow$
>           dynamic properties of transfer from r1 to r2

Notice that in these relationships, if an agent is allocated to a role, it might be assumed that the input and output ontologies of the agent are subsets of the role's input and output ontologies, but this assumption is not necessary. However, to satisfy the logical relationships specified above, at least a relevant overlap between the agent's ontologies and the role ontologies will be needed; for more discussion on this issue, see [41]. Moreover, note that if in the last relationship, A = B (an agent fulfilling two roles in one group), then dynamic properties of communication from A to A are required, i.e., that A will receive (at its input state) what it communicates (at its output state): 'A hears itself talking'. The logical relationships can be depicted as in the extension of Figure 3 shown as Figure 5. The following formalised criteria are required for an organisation realisation.

**Definition 9   (AGR Organisation Realisation)**
To realise an organisation, agents fulfil roles in the groups in the AGR organisation modelling approach. A realisation of an AGROrg organisation structure is then:
>   AGRReal = <AGROrg, Agents, fulfils>

A set of agent names is given, so Agents $\subseteq$ IDENT. The relationship
>       fulfils: Agents $\times$ Roles

specifies which roles an agent fulfils. The realisation dynamics specifies dynamic properties and ontologies for the agent input and output.
>    AGRRealDyn  =  < AGRReal , AGRDyn, agent_input_ontologies, agent_output_ontologies, agent_dynproperties>

where
>       agent_input_ontologies, agent_output_ontologies:         Agents $\rightarrow$ $\wp$(ONT)
>       agent_dynproperties:                        Agents $\rightarrow$ $\wp$(DYNPROP)

The agent must have the required ontologies on input and output, as well as the required properties must be implemented for the roles it fulfils:
>       $\forall$a $\in$ Agents, $\forall$r $\in$ Roles: fulfils(a, r) $\Rightarrow$

```
    role_input_ontologies(r) ⊆ agent_input_ontologies(a) ∧
    role_output_ontologies(r) ⊆ agent_output_ontologies(a) ∧    agent_dynproperties(a) ⊨
    role_dynproperties(r)
```

For all i ∈ Intergroup_interactions and for all a ∈ Agents, let involved_in(a, i) denote:

```
    ∃ r ∈ Roles:  fullfils(a, r) ∧ ( destination_of_interaction(r, i) ∨ source_of_interaction(r, i) )
```

The dynamical properties of intergroup interactions are assumed to be realised by having the same agent fulfil all roles in the intergroup interaction and having this agent process from its input to its output according to the intergroup interaction dynamical property; under this assumption, the intergroup interaction formally has to satisfy:

∀a ∈ Agents, i ∈ Intergroup_interactions:

```
    (∀r ∈ involved_roles(i): fulfils(a, r)) ⇒ agent_dynproperties(a) ⊨ inter-
    group_interaction_dynproperties(i)
```

The transfers also need to be realised by successfulness of sending messages by the involved agents:

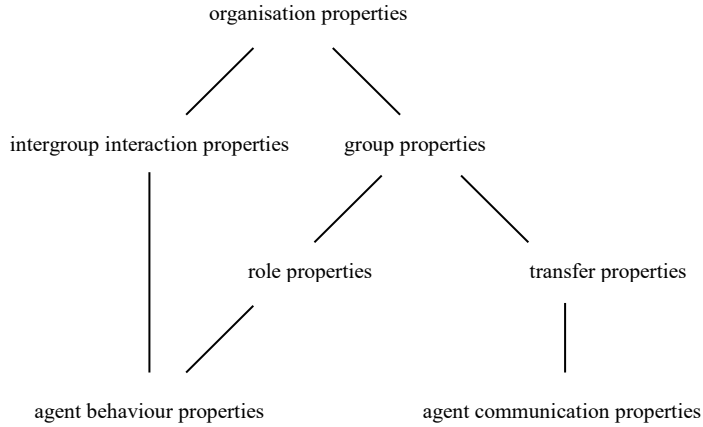

**Fig. 5.** Interlevel relations between dynamic properties for a realised organisation model

∀t ∈ Transfers, r1, r2 ∈ Roles:
```
    source_of_transfer(r1, t ) ∧ fullfils(a,r1) ∧
    destination_of_transfer(r2, t) ∧ fullfils(b,r2)  ⇒
    agent_commproperties(a, b)  ⊨ transfer_dynproperties(t)
```

For the example, the following allocation of agents agentA1, agentA2, agentB1, agentB2 to roles is possible:

| agentA1 - depA1 | agentB1 - depB1 | agentA1 - divArep |
| agentA2 - depA2 | agentB2 - depB2 | agentB1 - divBrep |

To realise the organisation model, for example agentA1 has to satisfy the following dynamic properties:

**DP(agentA1)**
    If        agent A1 receives progress information on component production,
    then     an updated planning will be generated by agent A1 taking this most recent

**IrRI(agentA1)**
    If        progress information on component production is received by agent A1,
    then      this will be communicated by agent A1 to agent B1
    If        agent A1 receives information on which components are needed,
    then      a component production planning will be generated by agent A1 taking
                these components into account.

Alternatively, if the roles in an intergroup interaction would not be fulfilled by one agent, but by several, this would create a mystery, since input to one agent creates output for another agent, even though the agents are not connected by any transfer since the roles they fulfil are from separate groups. This would suggest that the organisation structure is not complete. Therefore, in an AGR organisation model it is assumed that the roles in an intergroup interaction are fulfilled by one agent.

# 7 Relation to Organisational Structure From a Business Perspective

Organisational structure from a business perspective is to coordinate the activities of different people involved in a business, see e.g., [43]. People must now what their activity is and how it contributes to the product of the business and they should know what their role is in terms of responsibilities and to whom they are answerable. In this literature, the use of the word formal is different than in this paper. It recognises two types of organisation: the formal structure and the informal structure. In the formal structure three types of roles are distinguished:
- Line: each manager has authority over his subordinates,
- Staff: a group of advisers who do not have authority to command the general staff, but have the right and duty to advise managers, and
- Functional Authority: a manager or specialist that is given the authority to control the activities of people in more than one department.

The need for a Line structure can be explained from the concept of span of control. Span of control is the term for the number of subordinate employees directly accountable to a manager. The larger the number of employees a manager controls the wider is his span of control. If the span of control is too wide for the task at hand, then supervision becomes ineffective. It is because of span of control that a line structure has to develop.

    The formal structure defines the formal relationships of authority and subordination within a company using the three types of roles as a point of departure. In this context authority is defined by the right to make a decision, and responsibility as the duty to perform a task or activity that has been assigned. When a superior assigns a task to a subordinate the latter is given the authority (or right) to carry out the task, but he or she is also given the responsibility (duty) of carrying out the task. Authority is delegated from the top levels of the management down the organisation. Each position has rules governing what can and cannot be done. There are rewards and penalties for complying with these rules and performing duties well.

    The informal organisation refers to the emergent network of personal and social relations. The behaviour of people within these groups is governed by norms. When individuals break these norms, other members of the group impose sanctions on them. The effectiveness of the business not only depends on the functioning of the formal structure, but also of the informal structure.

    In large companies activities are further coordinated by organizing them into departments. Usually departments are organised by function or product, or combinations thereof. However, many other options exist.

This perspective on organisational structure relates to the work in this paper in the following way: the formal structure is modelled by our approach in terms of the roles in the structure. The three types of roles can be seen as blueprints for specific roles. For this the following relations are added to the formal framework:

    role_of_type: Roles x {line, staff, functional_authority}
    superior_of: Roles x Roles
    delegates_task_to: Roles x Tasks x Roles
    authorised_for: Roles x Tasks
    responsible_for: Roles x Tasks

Using this relation some global behaviour properties for roles of specific types can be provided, for example:

Line-property ≡ ∀r1, r2 ∈ Roles, t ∈ Tasks:
    role_of_type(r1, line) ∧ role_of_type(r2, line) ∧ superior_of(r1, r2) ∧ delegates_task_to(r1, t, r2) ∧
        authorised_for(r1, t) ∧ responsible_for(r1, t) ⇒
    authorised_for(r2, t) ∧ responsible_for(r2, t)

The rewards and penalties that belong to the formal structure are also specified as role behaviour for superiors towards subordinates in the Line structure. Furthermore, the above Line-property is can hold at the level of groups, but also holds for the whole organisation:

    Line-property ∈ organisation_dynproperties

The departments are formalized by groups, and the responsibilities of Staff and Functional Authority roles are specified as intergroup-interaction dynamic properties. The informal organisational structure as described in this section relates to the organisation realisation of Section 6. The social norms governing the behaviour of individual agents that fulfil roles of the formal organisation are specified in terms of generic agent dynamic properties and agent communication properties. Note that these properties typically cannot be designed along with the formal organisational structure. Therefore, by monitoring the functioning of the organisation realisation, i.e., the behaviour and communication of agents, and by applying pattern recognition techniques, these properties have to be detected as they evolve in the organisation.

# 7 Discussion

Both in human society and for complex systems of software agents, organisational structure is a means to make complex multi-agent dynamics manageable. To understand and formalize exactly how organisational structure relates to organisational dynamics is a fundamental challenge in the area of organisational modelling. The foundational framework relating organisational structure and dynamics as introduced in this paper provides support in addressing this challenge. Specification of organisation structure usually takes the form of pictorial descriptions, in a graph-like framework. These descriptions usually abstract from dynamics within an organisation. Specification of the dynamic properties of agent systems, on the other hand, usually takes place in a completely different conceptual framework; these dynamic properties are often specified in the form of a set of logical formulae in some temporal language. This paper puts forward a foundational metamodel for modelling languages that shows how these two perspectives can be related within one framework. It is shown how for different types of elements within an organisation structure model different sets of dynamic properties can be specified. The aspect of time in organisations can be studied formally using this framework, since the notion of time and dynamics is key to our approach. For example, the analysis done by Butler [45] can be formalised and where relevant simulated using our approach. The method can furthermore support the exploration of new organisational structures, for example the three models for knowledge-sharing in middle-sized companies of Mestad et al., [46]. By running simulations of formal organisation models and comparing those to the real (also informal) organisations, additional insights can be gained as to why a flexible model for organising knowledge-sharing seems to provide the greatest learning effect. In general, the combination of organisation modelling with an agent-based simulation approach is promising for the study of new institutional economics, see [47].

Illustrated for [12]'s AGR organisation modelling approach as a case study, it has been shown how a foundation can be obtained for modelling languages for integrated specification of both structure and dynamic properties of an organisation. The organisational structure provides structural relations between different elements of the organisation; these relations induce logical relationships between the sets of dynamic properties for the different elements of the organisation. From the perspective of meta-level reasoning and representation (e.g., [6], [33], [42]), these relationships between sets of properties, based on organisational structure, can be considered a metatheory, reasoning about a collection of object level theories (or viewpoints or local contexts) that represent the behavioural theories of the different elements of the organisation; cf. [2], [3], [5], [6], [20], [21], [42]. Such logical relationships make explicit dependencies between dynamic properties of parts of an organisation. The logical relationships express the kind of relations between dynamics of parts of an organisation, their interaction and dynamic properties of the organisation as a whole, which are indicated as crucial by (Lomi and Larsen, 2001) in their introduction.

How is this to be used in organisation (re)design or software system (re)design? Processes of organisational change (in succession) realise an evolving organisation or software system that is adapted to its (changing) environment; e.g., [31], [32]. From a strategic management or software engineering perspective, this evolution is to be guided: the organisational changes are to be initiated and accomplished. This usually will involve requirements on the dynamics of the whole system based on (the demands from) the environment, organisational goals, organisational structure, allocation of agents, and actions to achieve goals. In the approach presented in this paper, to obtain a changed organisational structure is just a means to achieve requirements on organisational dynamics. Such requirements concerns (dynamic) properties of the organisational behaviour. So, the (re)design process starts by identifying the requirements in terms of the required changes in organisational behaviour, given changed environmental conditions; see the right hand side in Figure 6.

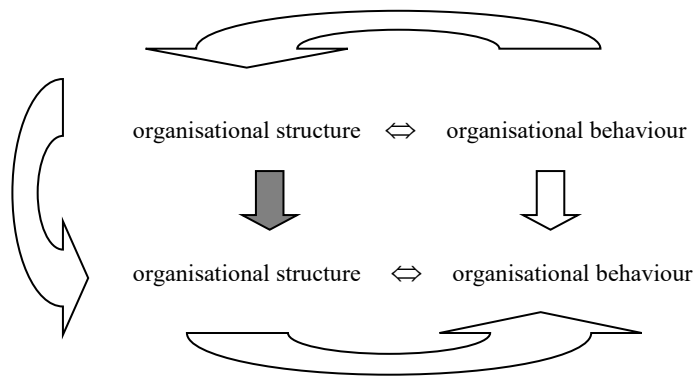

**Fig. 6** Organisational (re)design of a multi-agent system

Next, using the relations discussed in this paper (indicated by the horizontal ⇔ arrows) between organisational behaviour and organisational structure and allocation of agents, it is determined which changes in organisational structure and allocations of agents (the left hand side of Figure 6) are needed to achieve the required changes in organisational behaviour. Finally, the changes in organisational structure and allocation of agents are actually made. Thus what is achieved is a change in organisational behaviour; see the right downward arrow in Figure 6. This behavioural change is realised by means of a change in organisation structure and allocation of agents: the left downward arrow.

The metamodel presented here contributes directly to the foundations of some of the criteria proposed by [9] (e.g., the criteria regarding the identification of constraints and flexible analysis) for an ideal social simulation language (ISSL), and it illustrates for some of the other criteria how such an ISSL can come about (e.g., the criteria regarding modeller's intentions, compositionality, and emergence) in the context of the structure and dynamics of organizations. Furthermore, a shared basis of the work of [9] and the work presented here lies in the view that the different simulation processes need to be specified separately from the significant outcomes of the simulation, and that the processes that emerge from the simulation are typically more interesting than the resulting states of the simulation.

Also in MOISE and MOISE+ (cf. [16], [23], [26]), as for AGR, organisational structures are based on roles and groups and on links between roles. A difference with AGR is that tasks are taken into account explicitly and specified by goals, plans, resources and actions. These specifications can be considered one of the specific instantiations possible for the notion of dynamic property that is central in our foundational perspective as developed. In this case interlevel relations take the form of relations between tasks for roles and tasks at a higher aggregation level, for example for groups or (parts of) the organisation.

Having one framework that integrates the two perspectives, as well as the logical relationships between the two perspectives enables formal diagnostic analysis. Any simulation or empirical trace can be checked against a given dynamic property. Assuming the logical relationships, diagnosis of dysfunctioning within an organisation can be performed see, for example, [27].

For further work, this metamodel provides a basis for development of more specific organisation specification languages, covering both structure and dynamics. Such languages could make use of semiformal and graphical elements, and specification would be supported by software tools. The specification tools can mediate between a user and further software environments for simulation and analysis. For example, the work of Aken [44] can be combined with our approach by simulating his process model of organisational change and applying it to our formal models of organisations. Moreover, verification techniques and tools (e.g., [9], [16], [33]) can be considered for verification of dynamic properties and their interlevel relations.

## Acknowledgements

The authors are grateful to Wouter Wijngaards who has contributed to some elements of this paper, in particular in Section 3, and to two anonymous referees who pointed out how the paper could be improved in its motivation and explanation.